\documentclass[10pt,twocolumn,letterpaper]{article}

\usepackage{wacv}              

\usepackage{graphicx}
\usepackage{amsmath}
\usepackage{amssymb}
\usepackage{booktabs}

\usepackage{xcolor}
\usepackage{tikz}
\usetikzlibrary{arrows}
\usetikzlibrary{arrows.meta}
\usepackage{dsfont}

\DeclareMathOperator*{\argmax}{arg\, max}

\definecolor{blizzardblue}{rgb}{0.67, 0.9, 0.93}
\definecolor{blond}{rgb}{0.98, 0.94, 0.75}
\definecolor{bubblegum}{rgb}{0.99, 0.76, 0.8}

\usepackage[pagebackref,breaklinks,colorlinks]{hyperref}

\usepackage[capitalize]{cleveref}
\crefname{section}{Sec.}{Secs.}
\Crefname{section}{Section}{Sections}
\Crefname{table}{Table}{Tables}
\crefname{table}{Tab.}{Tabs.}

\def\wacvPaperID{1427} 
\def\confName{WACV}
\def\confYear{2024}

\begin{document}

\title{Uncertainty-weighted Loss Functions for Improved Adversarial Attacks on Semantic Segmentation}

\author{Kira Maag\\
Technical University of Berlin, Germany\\
{\tt\small maag@tu-berlin.de}
\and
Asja Fischer\\
Ruhr University Bochum, Germany\\
{\tt\small asja.fischer@rub.de}
}
\maketitle

\begin{abstract}
State-of-the-art deep neural networks have been shown to be extremely powerful in a variety of perceptual tasks like semantic segmentation. However, these networks are vulnerable to adversarial perturbations of the input which are imperceptible for humans but lead to incorrect predictions. Treating image segmentation as a sum of pixel-wise classifications, adversarial attacks developed for classification models were shown to be applicable to segmentation models as well. In this work, we present simple uncertainty-based weighting schemes for the loss functions of such attacks that (i) put higher weights on pixel classifications which can more easily perturbed and (ii) zero-out the pixel-wise losses corresponding to those pixels that are already confidently misclassified. The weighting schemes can be easily integrated into the loss function of a range of well-known adversarial attackers with minimal additional computational overhead, but lead to significant improved perturbation performance, as we demonstrate in our empirical analysis on several datasets and models. 
\end{abstract}

\section{Introduction}
\begin{figure}
    \centering
    \begin{subfigure}[t]{.48\linewidth}
    \centering\includegraphics[trim=0 0 2048 512,clip,width=0.99\textwidth]{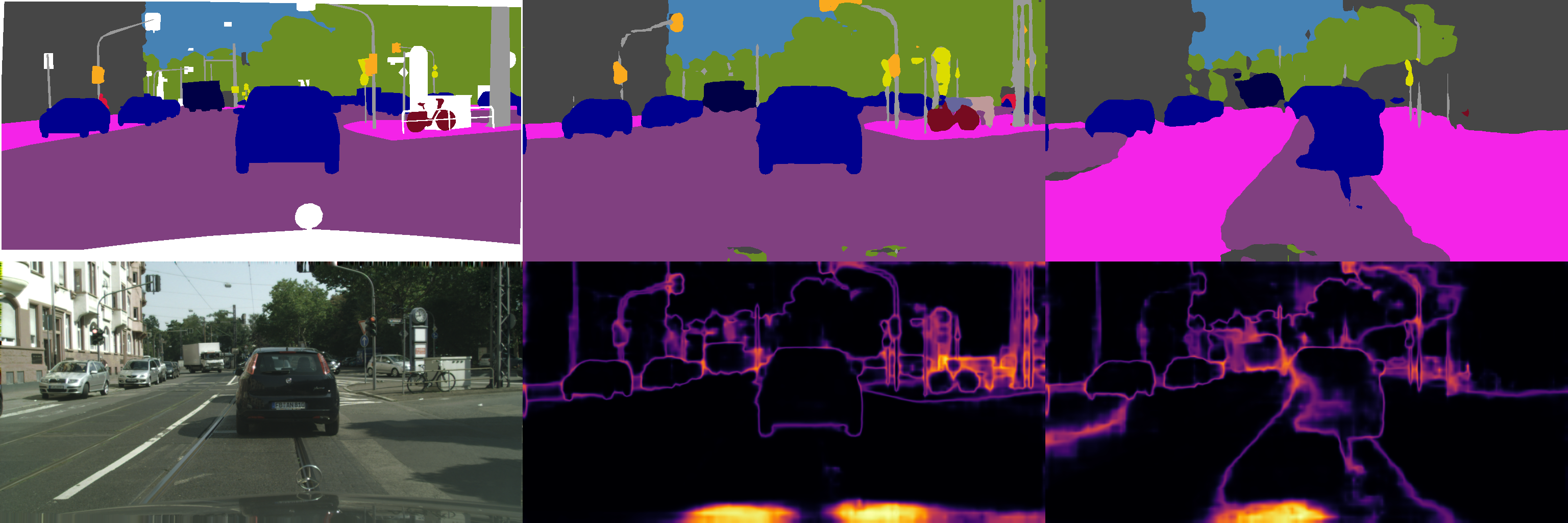}
    \caption{Input image}
    \end{subfigure}
    \begin{subfigure}[t]{.48\linewidth}
    \centering\includegraphics[trim=1024 512 1024 0,clip,width=0.99\textwidth]{figs/frankfurt_000000_003025.png}
    \caption{Clean}
    \end{subfigure}
    \\
    \begin{subfigure}[t]{.48\linewidth}
    \centering\includegraphics[trim=2048 512 0 0,clip,width=0.99\textwidth]{figs/frankfurt_000000_003025.png}
    \caption{Perturbed by original attack}
    \end{subfigure}
    \begin{subfigure}[t]{.48\linewidth}
    \centering\includegraphics[trim=2048 512 0 0,clip,width=0.99\textwidth]{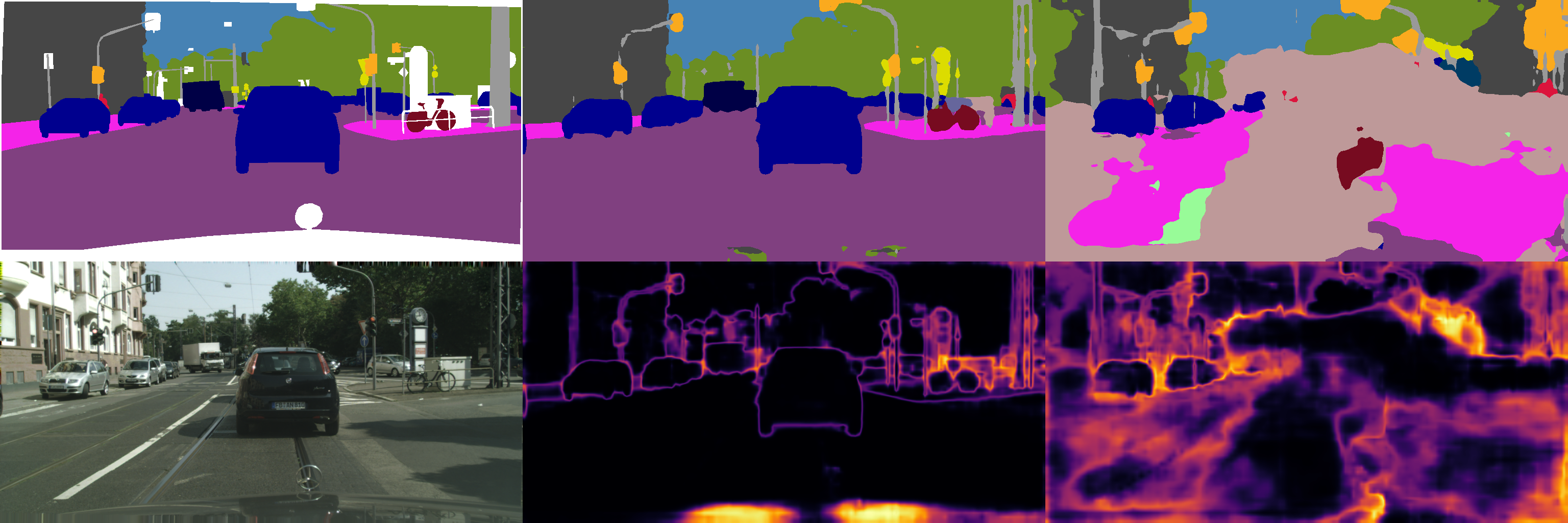}
    \caption{Perturbed by our method}
    \end{subfigure}
    \caption{Semantic segmentation prediction for a clean image, a perturbed image generated by the original attack proposed in \cite{Kurakin2017} and a perturbed image created by the same attack using our uncertainty-weighted loss.}
    \label{fig:pred_example}
\end{figure}

Deep neural networks (DNNs) have been shown to be extremely powerful in a wide range of perceptual tasks, such as semantic image segmentation \cite{Chen2018,Pan2022} for which they demonstrate an outstanding prediction performance. Semantic segmentation provides comprehensive and precise information about the given image by assigning each pixel to a predefined and fixed set of semantic classes resulting in segmented objects.
However, many studies have found that DNNs are vulnerable to \emph{adversarial attacks} \cite{Arnab2018,Bar2021}. Adversarial attacks generate slightly perturbed versions of the input images which fool the DNN, i.e., change the network predictions at test time, see for example \cref{fig:pred_example}. These small perturbations are not perceptible to humans making adversarial examples very hazardous in safety-related applications like automated driving. 
Thus, the development of efficient defense strategies against adversarial attacks is of highest interest. These strategies either enhance the robustness of DNNs rendering it more challenging to generate adversarial examples, or rely on approaches to detect adversarial attacks. 
In general, there are three common approaches to increase the robustness of DNNs. 
The first class  of approaches aims to enhance the robustness of the network by modifying the training process (e.g.\cite{Klingner2020,Xu2021}). Second, input denoising procedures, like autoencoder based reconstruction \cite{Cho2020} or inpainting \cite{Yatsura2022}, are considered to remove the perturbation from the input.
The third class of approaches increases the robustness during inference, e.g.\ by multi-scale processing \cite{Arnab2018}.
For the detection of adversarial examples, the patch-wise spatial consistency check \cite{Xiao2018} and an uncertainty-based method \cite{Maag2023} have been proposed.
Regardless of strategy, it is a continuous loop between the development of defense/detection strategies and adversarial attackers. This also means that the development of new faster and stronger attacks is important in order to strengthen the models against them and thus, enhancing general model robustness. 

Prior works on adversarial attacks focus on the image classification task and some of them are transferred to the semantic segmentation task by treating each pixel labeling independently as a separate classification task \cite{Goodfellow2015,Kurakin2017,Madry2018}. Moreover, there were adversarial examples specifically developed for the semantic segmentation task where all pixels of an image are attacked until selected pixels have been misclassified into the target class, i.e., pixels of a selected class appear or disappear, or even the entire image changes \cite{Cisse2017,Metzen2017}. In comparison to these methods, the patch-wise attack \cite{Nakka2020,Nesti2022} perturbs a small rectangular region of the image aiming to cause prediction errors in the whole image. Recently, the certified radius-guided attack framework for segmentation models \cite{Qu2023} was introduced. The idea is to disturb pixels with comparatively smaller certified radii since a smaller theoretically certified radius should relate to lower robustness to adversarial perturbations.

In this paper, we present an uncertainty-based weighting scheme which can be incorporated into the loss function of any untargeted attack on semantic segmentation models that is composed out of pixel-wise attacks. 
Uncertainty information, such as Monte-Carlo Dropout \cite{Gal2016} or maximum softmax \cite{Hoebel2020}, is considered for prediction error \cite{Maag2019,Wickstrom2018} and out-of-distribution detection \cite{Maag2023_grads}. These works demonstrate the correlation between uncertainty measures and erroneous predictions. 
The idea behind our method is to include uncertainty information into the loss function of an adversarial attack to degrade the performance of the network even more. To this end, we consider different \emph{white box} attacks, i.e., the attacker has full access to the model including parameters and loss function used during training. In contrast, \emph{black box} methods gain zero knowledge about the model to attack. 
The expectations for any attack are low runtimes and computational effort while at the same time having powerful perturbation effects. We modify the loss function of well-known adversarial methods by introducing an uncertainty-weighted loss. On the one hand, we put higher weights on pixel classifications which can more easily perturbed and on the other hand, we zero-out the pixel-wise losses corresponding to those pixels that are already confidently misclassified.
Thus, our approach can be incorporated into any attack with minimal additional computational overhead, but improved perturbation performance. 
First, we apply our loss function to pixel-wise attacks for semantic segmentation.
Second, we replace the certified radius-guided loss function introduced in \cite{Qu2023} by our uncertainty-based weighting scheme.
Last, we introduce an alternative approach for the patch-based attack where only a few pixels are attacked. To this end, we choose randomly a subset of pixels to attack and apply our loss function in combination with the pixel-wise iterative \emph{fast gradient sign method} \cite{Kurakin2017}. 

In our tests, we employ state-of-the-art semantic segmentation networks \cite{Chen2018,Pan2022,Yu2018,Zhao2017} applied to the Cityscapes \cite{Cordts2016} as well as the Pascal VOC2012 dataset \cite{Everingham2012} demonstrating our adversarial attack performance. We apply our approach to different types of attacks, such as pixel-level attackers designed for image classification \cite{Goodfellow2015,Madry2018} and pixel-wise attacks developed for semantic segmentation \cite{Rony2022,Qu2023}. 
The source code of our method is publicly available at 
\url{https://github.com/kmaag/Uncertainty-weighted-Loss}.
Our contributions are summarized as follows:
\begin{itemize}
    \item For the first time, we present an uncertainty-based weighting scheme which can be incorporated into the loss function for white box adversarial attacks which has low computationally overhead compared to the original attack.
    \item Our method is not designed for a specific adversarial attack, rather we enhance different types of attackers in a light-weight manner. We achieve attack pixel success rate values of up to $99.82\%$ across different network architectures and datasets.
    \item We propose an approach which attacks only a subset of the image pixels, similar to the patch attack, but also leading to erroneous prediction of the entire image.
\end{itemize}
The paper is structured as follows. In \cref{sec:background}, we present various adversarial attacks for the semantic segmentation task which serve as baselines in our work. We introduce our method in \cref{sec:method}. In \cref{sec:exp}, the numerical results are shown, followed by a comparison with related work in \cref{sec:rel_work} and a conclusion in \cref{sec:conc}.
%
%
%
\section{Background}\label{sec:background}
In this section, we recall the semantic segmentation task as well as different previously proposed adversarial attackers to semantic segmentation models. Note, that all described attacks belong to the white box setting.
%
%
\paragraph{Semantic segmentation}
To obtain a semantic segmentation, i.e., pixel-wise classification of image content, each pixel $z$ of an input image $x$ gets assigned a label $\tilde y_z$ from a prescribed label space $C = \{y_{1}, \ldots, y_{c} \}$. A neural network given learned weights $w$ provides for the $z$-th pixel a probability distribution $f(x;w)_{z} \in \mathbb{R}^{|C|}$ specifying the probability for each class $y \in C$ denoted by $p(\cdot|x)_z \in \mathbb{R}$. 
The predicted class is then computed by $\hat y_{z}^x =\argmax_{y\in\mathcal{C}} p(y|x)_z$.
To train the semantic segmentation network, a pixel-wise loss function (generally the cross entropy) is simultaneously minimized for all pixels $z \in Z$ of an image $x$. The complete loss function is then given by
\begin{equation}\label{eq:loss}
    L(f(x;w),y) = \frac{1}{|Z|} \sum_{z \in Z} L_{z}(f(x;w)_{z},y_{z})  \enspace,
\end{equation} 
where $y_{z}$ denotes the one-hot vector of the label.
%
%
\paragraph{Adversarial attacks}
A well-known adversarial attack developed for image classification but also applied to semantic segmentation is the \emph{fast gradient sign method} (FGSM, \cite{Goodfellow2015}).
This (untargeted) single-step attack adds small perturbations to the image $x$ leading to an increase of the loss of 
\begin{equation} \label{eq:fgsm}
x^{\mathit{adv}} = x + \varepsilon \cdot \text{sign}(\nabla_x L(f(x;w),y)) \enspace, 
\end{equation}
where $\varepsilon$ describes the magnitude of perturbation, i.e., the $\ell_{\infty}$-norm of the perturbation is bounded to be (at most) $\varepsilon$. 
This attack is extended to the \emph{iterative FGSM} (I-FGSM, \cite{Kurakin2017}) increasing the perturbation strength by 
\begin{equation} \label{eq:ifgsm}
    x^{\mathit{adv}}_{t+1} = \text{clip}_{x,\varepsilon} (x^{\mathit{adv}}_{t} + \alpha \cdot \text{sign}(\nabla_{x^{\mathit{adv}}_{t}} L(f(x^{\mathit{adv}}_{t};w),y))) \enspace,
\end{equation}
where $x^{\mathit{adv}}_{0} = x$, $\alpha$ defines the step size, and a clip function ensures that $x^{\mathit{adv}}_{t} \in [x-\varepsilon, x+\varepsilon]$. 
The \emph{projected gradient descent} (PGD, \cite{Madry2018}) attack is similar to the iterative FGSM. The difference between the two methods is that PGD choose the starting point randomly within the $\ell_{\infty}$ ball of interest (and does random restarts), while I-FGSM initializes to the original point.
These approaches serve as basis for further elaborated attacks like the \emph{orthogonal PGD} \cite{Bryniarski2022} or \emph{DeepFool} \cite{Moosavi2016}.

Furthermore, there have been developed adversarial attacks especially for the semantic segmentation task such as adaptions of the PGD attack \cite{Agnihotri2023,Gu2022}. 
The introduced \emph{ALMA prox} attack \cite{Rony2022} is based on a proximal splitting to produce adversarial perturbations with much smaller $\ell_{\infty}$-norm in comparison to FGSM and PGD.
Recently, a \emph{certified radius-guided} (CR) attack framework for segmentation models was proposed \cite{Qu2023}. The certified radius specifies the size of an $\ell_p$ ball around a pixel in which a perturbation is guaranteed to not change the class predicted for the pixel. 
Thus, a larger certified radius indicates more robustness to adversarial perturbations. The idea of the framework is to focus on disrupting pixels with relatively smaller certified radii.

In contrast to attacks adding perturbations to all pixels, \emph{patch attacks} \cite{Nakka2020} disrupt a small rectangular region of the image aiming at prediction errors
in a much larger region, i.e., the whole image. 
In \cite{Nesti2022}, an individual patch attack is introduced, the \emph{expectation over transformation-based attack}, creating robust adversarial examples to perturb a range of transformations at the same time. In the real world scenario, transformations consist of angle and viewpoint changes for instance. To generate these strong perturbing patches for the semantic segmentation task within the optimization procedure an extension of the pixel-wise cross entropy loss is introduced.

There also exist targeted attacks specifically developed for semantic segmentation. 
For the \emph{stationary segmentation mask method} \cite{Cisse2017,Metzen2017,Xie2017} the pixels of an image are iteratively perturbed until most of the pixels have been misclassified as belonging to the target class given by an arbitrary segmentation defined by the attacker. 
The \emph{dynamic nearest neighbor method} \cite{Chen2022,Metzen2017} is intended to remove one desired target class (like pedestrians from street scene images) but keep for all other classes the network’s segmentation unchanged.
%
%
%
\section{Uncertainty-weighted loss functions}\label{sec:method} 
\begin{figure}
    \centering
    \includegraphics[width=0.36\textwidth]{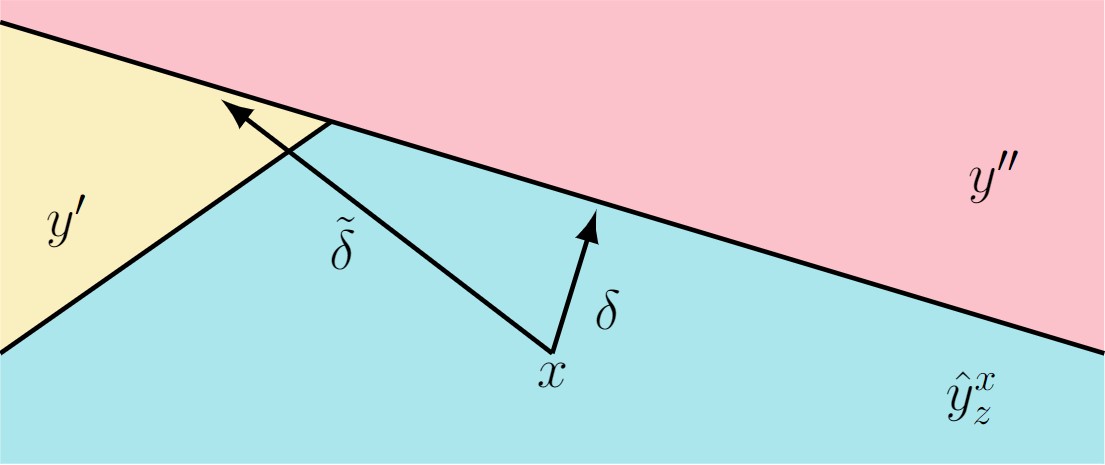}
    \caption{An illustration of the decision boundaries between three different classes where $\hat{y}^x_z$ is the predicted class of input $x$, $y''$ the class with second highest probability and $y'$ another class. The arrows $\delta$ and $\tilde{\delta}$ represent distances from $x$ to the  decision boundaries.}
    \label{fig:idea}
\end{figure}
In the following, we describe the two weighting schemes which follow different motivations: on the one hand focusing on pixels whose classification is easily perturbable, and on the other paying no attention to those pixels which are already misclassified with high confidence.
%
%
\subsection{Focusing on easily to perturb pixels}\label{sec:method_unc} 
Recall, that for creating an adversarial attack we want to add a perturbation (of a small predefined magnitude) to the image that changes as many pixel-wise classifications as possible. Intuitively, due to the restriction of the magnitude of the perturbation, it makes sense to focus on those pixel classifications that are easily to perturb. Geometrically, a small shift of the input is more likely to lead to a different classification result as closer the input is to the decision boundary between the current and other classes.
Let $\hat y_z^x$ be the predicted class for pixel $z$ and lets define $y'$ to be another class. 
In a linear model the minimal distance to the decision boundary separating class $\hat y_z^x$ from class $y'$ is then defined by 
\begin{equation}\label{eq:delta}
    \delta = \frac{p(\hat y^x_{z}|x)_z - p(y'|x)_z}{|| \nabla_x (p(\hat{y}^x_{z}|x)_z - p(y'|x)_z) ||_2} \enspace.
\end{equation}
An illustration is given in \cref{fig:idea}. 
If an attack would focus on the misclassification of only a single pixel it would make sense to focus at the one with the smallest distance to the second most likely class. In this case the numerator of \cref{eq:delta} is given by the probability margin
\begin{equation}
    M(x)_{z} = p(\hat y_{z}^x|x)_z - \argmax_{y\in\mathcal{C}\backslash \{ \hat y_z^x \}} p(y|x)_z \enspace.
\end{equation}
However, neural networks are not linear models and the attacks aim at the misclassification of all pixels. 
Therefore, the adversarial perturbation of the input does not necessarily points into the  direction of shortest distance to a decision boundary. To take this into account a better indication of which pixel classification is easiest to attack could be given by the difference between the highest and lowest class probability  
\begin{equation}
    D(x)_{z} = p(\hat y_{z}^x|x)_z - \min_{y\in\mathcal{C}} p(y|x)_z \enspace, 
\end{equation}
which serves as a proxy of maximal distance to a decision boundary, i.e., the distance to the least likely class.
If the prediction is highly certain, the probability of the least likely class is equal or close to zero. 

Instead, one can also estimate the mean of the margins to all other classes
\begin{equation}
    \bar{M}(x)_{z} = \frac{1}{|C|-1} \sum_{y\in\mathcal{C}\backslash \{ \hat y_z^x \}} p(\hat y_{z}^x|x)_z - p(y|x)_z \enspace.
\end{equation}
Moreover, the entropy 
\begin{equation}
    E(x)_{z} =-\sum_{y\in \mathcal{C}}p(y|x)_{z} \cdot \log p(y|x)_{z} 
\end{equation}
could be a good indicator, since it is often used as uncertainty measure for the semantic segmentation task and it is related to a weighted mean margin (where each margin is weighted by $log(p(y|x)_{z})$, i.e., smaller margins get a higher weight).

The proposed indicators for the closeness to decision boundaries can than be used as weighting factors for the pixel-wise loss functions when calculating the adversarial attacks. Such weighted versions of the FGSM and I-FGSM method would for example be obtained by replacing  $L(f(x;w),y)$ in \cref{eq:fgsm} and \cref{eq:ifgsm}, respectively, by
\begin{equation}\label{eq:loss_uw}
    L^{U}(f(x;w),y) = \frac{1}{|Z|} \sum_{z \in Z} e^{U(x)_z} \cdot L_{z}(f(x;w)_{z},y_{z})
\end{equation}
where $U(x)_z \in \{ 1-M(x)_z, 1-D(x)_z, 1-\bar M(x)_z,  E(x)_z \} $.
The pixels with high uncertainty, i.e., larger values of $U(x)_z$, are weighted stronger in the loss during the adversarial example generation in order to lead as many pixels as possible to a wrong prediction. We re-scale the uncertainty values $U(x)$ by the exponential function to further emphasize high uncertainties.
%
%
\subsection{Ignoring confidently wrongly classified pixels}
If pixels are already predicted incorrectly, it makes no sense to continue to give them a high weighting. Thus, it is important to focus on pixels, when applying the loss function, that are still correctly classified. 
That is, the loss of pixels which are already misclassified with sufficiently high confidence can be neglected. Therefore, during the attack generation process, we set the loss of all pixels which are misclassified with a probability of at least $75\%$ to zero. This is achieved by the following weighting scheme
\begin{align}\label{eq:loss_zero}
    L^{0}(f(x;w),y) = & \frac{1}{|Z|} \sum_{z \in Z} 
    \mathds{1}_{ (\hat{y}_z^x = \tilde{y}_z \lor p(\hat{y}_z^x |x) < 0.75)}  \nonumber \\
    & \cdot L_{z}(f(x;w)_{z},y_{z}) 
\end{align}
with ground truth class $\tilde{y}_{z}$. 
The confidence of the misclassification (as measured by $p(\hat{y}_z^x |x)$) has to be taken into account, since the corresponding pixel is still perturbed based on the gradients of the other pixel-wise loss functions. If the uncertainty is high, i.e., the confidence is low, this could shift the pixel accidentally  back into the correct class.  
In general, our uncertainty-based weighting scheme, $L^U$ and $L^0$, can be inserted into the loss function used in any untargeted and pixel-wise adversarial attack.
%
%
%
\section{Experiments}\label{sec:exp}
In this section, we describe the experimental setting first and then evaluate our adversarial attack performance.
\begin{figure*}
    \centering\includegraphics[width=0.98\textwidth]{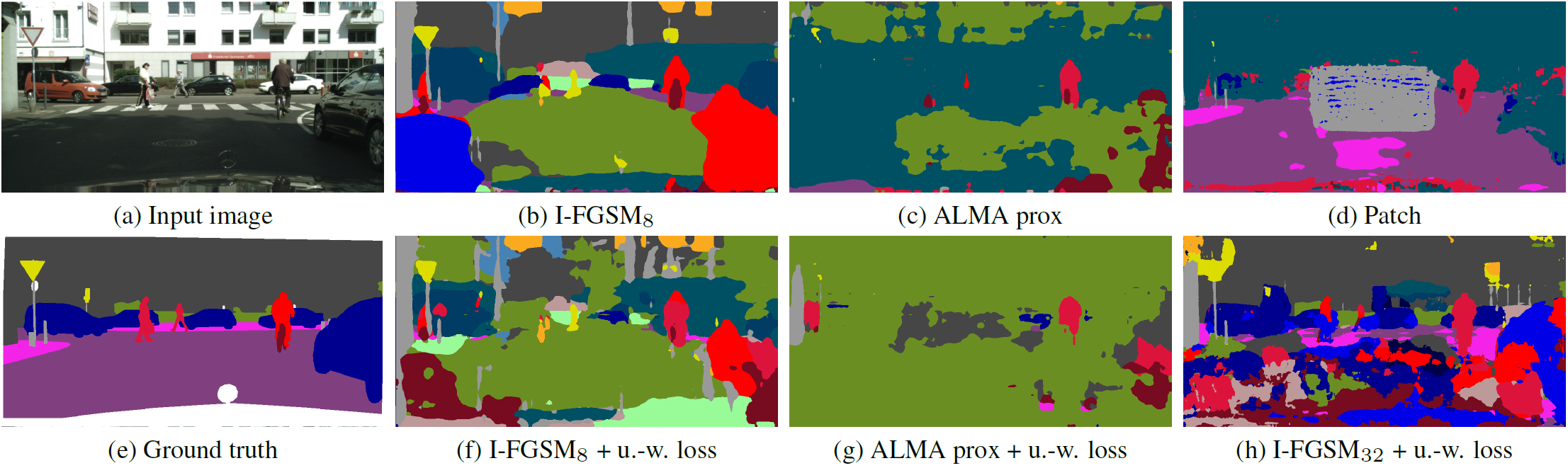}
    \caption{Segmentations for weighted and un-weighted attacks. (a) Input image from the Cityscapes dataset and (e) corresponding ground truth. Semantic segmentation prediction for perturbed images generated by (b) iterative  FGSM, (c) ALMA prox, (d) patch attack, (f) FGSM with uncertainty-weighted (u.-w.) loss, (g) ALMA prox with u.-w. loss and (h) iterative FGSM with u.-w. loss applied to only a subset of pixels.}
    \label{fig:pred_example_diff}
\end{figure*}
\begin{figure}
    \centering
    \includegraphics[width=0.48\textwidth]{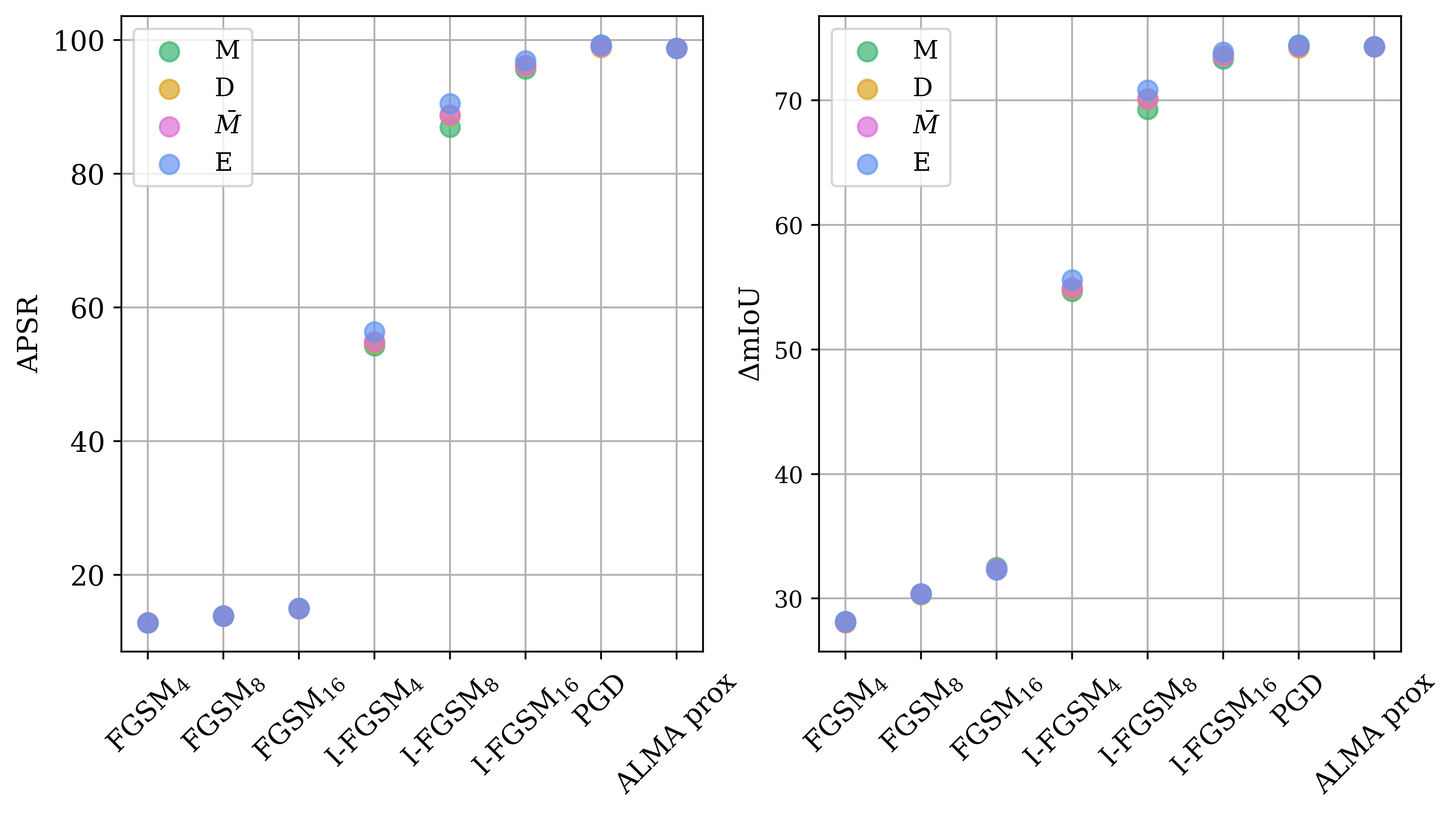}
    \caption{Comparison of uncertainty measures $M$, $D$, $\bar M$ and $E$ used for weighting loss functions of different adversarial attacks for the DeepLabv3+ network applied to the Cityscapes dataset.}
    \label{fig:unc_measures}
\end{figure}
%
%
%
\subsection{Experimental setting}\label{sec:exp_setting}
%
%
\paragraph{Datasets}
The experiments are conducted on two datasets, Cityscapes \cite{Cordts2016} and Pascal VOC2012 \cite{Everingham2012} (shorthand VOC). The latter dataset, for visual object classes in realistic scenes, consists of $1,\!464$ training and $1,\!449$ validation images with annotations for different objects of categories person, animal, vehicle and indoor. The Cityscapes dataset, for semantic segmentation in street scenes, contains $2,\!975$ training and $500$ validation images of dense urban traffic in $18$ and $3$ different German towns, respectively. 
%
%
\paragraph{Segmentation networks}
In our tests, we consider four different pre-trained state-of-the-art networks. Trained on the Cityscapes dataset, the BiSeNet \cite{Yu2018} achieves a mean intersection over union (mIoU) of $74.37\%$ on the validation set and the DDRNet \cite{Pan2022} of $77.80\%$. Moreover, we employ the DeepLabv3+ network \cite{Chen2018} trained on Cityscapes obtaining a validation mIoU of $79.61\%$ and on VOC of  $76.81\%$. The PSPNet \cite{Zhao2017} achieves a mIoU value of $76.78\%$ on the VOC validation set.
%
%
\paragraph{Adversarial attacks} 
We consider the well-known and in defense approaches often considered \cite{Arnab2018,Bar2021,Cho2020,Klingner2020} FGSM and I-FGSM attacks in our tests with parameter setting proposed in \cite{Kurakin2017}. The step size is given by $\alpha=1$ and the perturbation magnitude by $\varepsilon = \{ 4,8,16 \}$ resulting in a number of iterations of $n = \min \{ \varepsilon+4, \lfloor 1.25\varepsilon \rfloor \} $.
The corresponding (iterative) FGSM attack is denoted by FGSM$_{\varepsilon}$ and I-FGSM$_{\varepsilon}$. 
As another pixel-wise attack originally developed for image classification, we use 
the PGD attack with parameters $\alpha=1/30$, $\varepsilon=1$, $n=40$ and one restart. 
In addition, we employ various adversarial example generation techniques specifically designed for the semantic segmentation task. Firstly, we consider the ALMA prox attack with default parameters using the implementation of \cite{Rony2022}. For the described attacks above, we employ the model zoo\footnote{\url{https://github.com/open-mmlab/mmsegmentation}} including the pre-trained models.
Secondly, for the certified radius-guided approach \cite{Qu2023}, we use the provided code with two parameter settings, i.e., $\ell_{2}$-norm with $\varepsilon=1$ and $\ell_{\infty}$-norm with $\varepsilon=0.004$ (hyperameters all under default setting).
Lastly, we consider the patch attack \cite{Nesti2022} using the available repository with default parameters applied to the BiSeNet and the DDRNet tested on the real world Cityscapes dataset.

The model used to compute the certified radius-guided method re-scales the VOC images to $473 \times 473$ which we keep also for the other attacks.
Since the Cityscapes dataset provides high-resolution images of size $1024 \times 2048$, we re-scale the image size to $512 \times 1024$ for the computation of the adversarial examples to reduce the amount of memory to run a full backward pass. 
\Cref{fig:pred_example_diff} (top row) shows semantic segmentation predictions for a few attacks applied to the Cityscapes dataset and the BiSeNet network.
%
%
\paragraph{Evaluation metrics}
To access the performance of the adversarial attackers, we use the attack pixel success rate (APSR) \cite{Rony2022} which is defined by
\begin{equation}\label{eq:apsr}
    \text{APSR} = \frac{1}{|Z|} \sum_{z \in Z} \argmax_{y\in\mathcal{C}} p(y|x)_z \neq \tilde{y}_{z}
\end{equation}
with ground truth class $\tilde{y}_{z}$. This metric measures the number of falsely predicted pixels and thus, successfully attacked pixels.
Furthermore, we consider the difference of the mIoU obtained on clean images and the mIoU obtained on perturbed images as performance metric, denoted by $\Delta$mIoU. Note, this metric is bounded by the mIoU value on clean images.
%
%
\subsection{Comparison of different weighting schemes}
In \cref{sec:method_unc}, we introduced four different measures for estimating which pixel classification can easily be disturbed: the probability margin $M$, the difference between the highest and lowest probability value $D$, the mean of the margins $\bar M$, and the entropy $E$. 
The first experiment aimed at analysing which of these measures performs best. For that, we have computed the different weighted attacks on the DeepLabv3+ network applied to the Cityscapes dataset. \Cref{fig:unc_measures} shows a comparison of the performance in terms of APSR and $\Delta$mIoU.
The scores are fairly close and for the weaker (FGSM) as well as stronger attacks (PGD and ALMA prox) all measures lead to almost the same results. For all attacks, the values resulting from weighting with the difference between the highest and lowest probability $D$ and the mean of the margins $\bar{M}$ are almost equal. Overall, the probability margin performs the worst, while the entropy shows slight improvements for the iterative FGSM attack. These results indicate, that it is advantageous to consider the margin for more than one class and that higher weighting margins to more likely classes (as approximately done by the entropy) is also beneficial. 
Thus, for the following experiments, we consider only the entropy as uncertainty measure.
%
%
\subsection{Evaluation of uncertainty-weighted attacks}\label{sec:results_loss}
\begin{figure*}
    \begin{subfigure}[t]{.5\linewidth}
    \centering\includegraphics[width=0.99\textwidth]{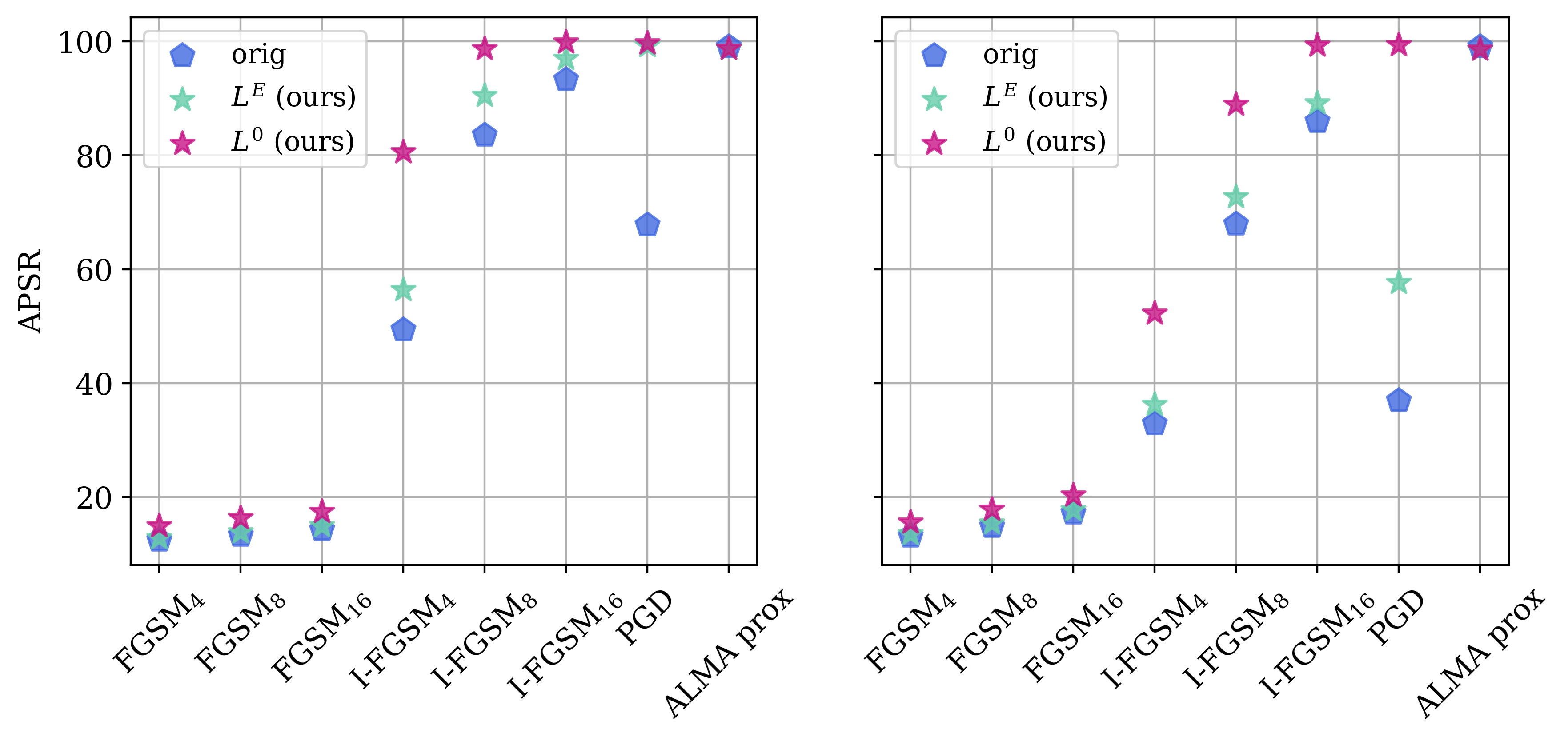}
    \caption{DeepLabv3+ (left) and BiSeNet (right)}
    \end{subfigure}
    \begin{subfigure}[t]{.5\linewidth}
    \centering\includegraphics[width=0.99\textwidth]{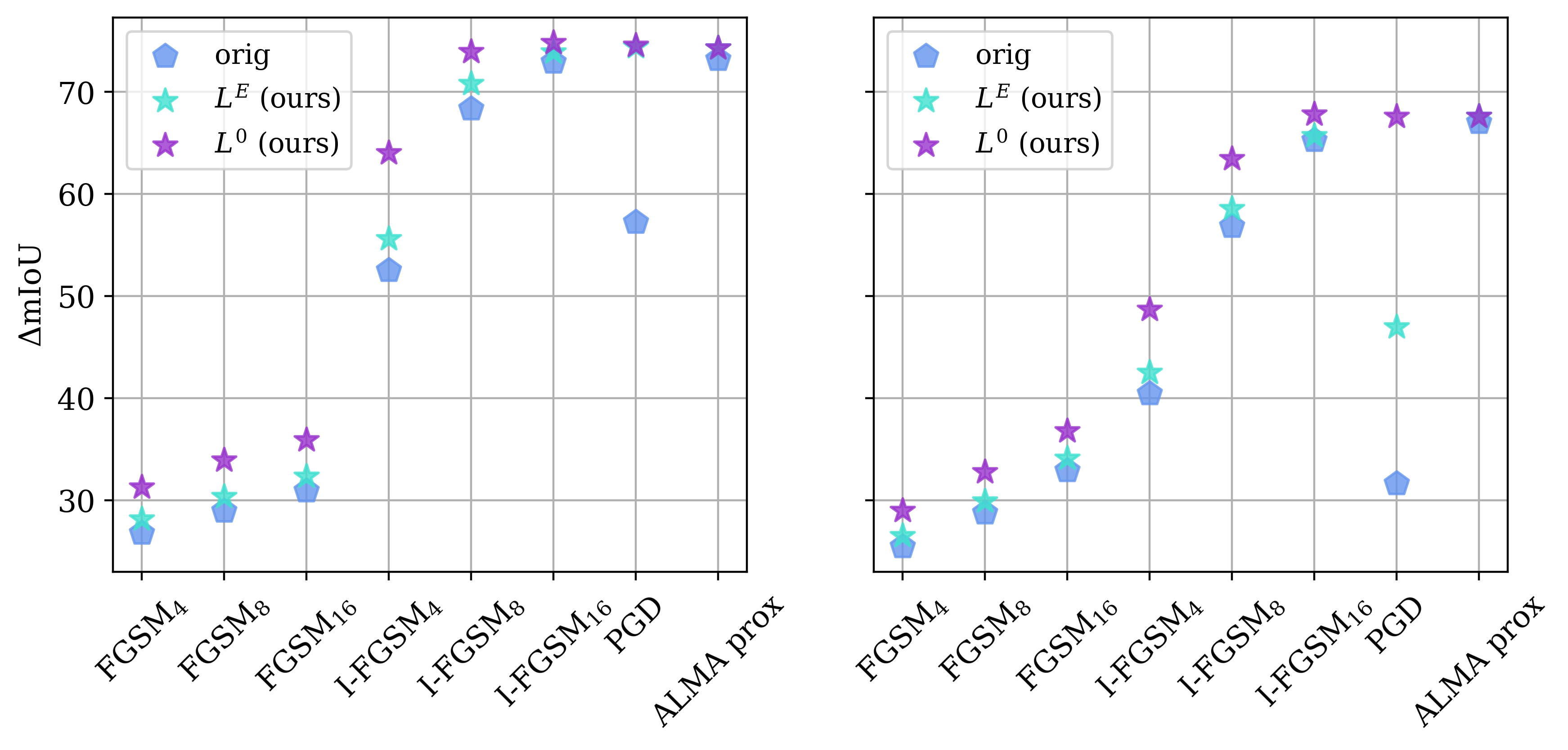}
    \caption{DeepLabv3+ (left) and BiSeNet (right)}
    \end{subfigure}
    \caption{APSR (a) and $\Delta$mIoU (b) results for different attacks on two networks trained on the Cityscapes dataset.}
    \label{fig:apsr_miou_cs}
\end{figure*}
\begin{figure*}
    \begin{subfigure}[t]{.5\linewidth}
    
    \centering\includegraphics[width=0.99\textwidth]{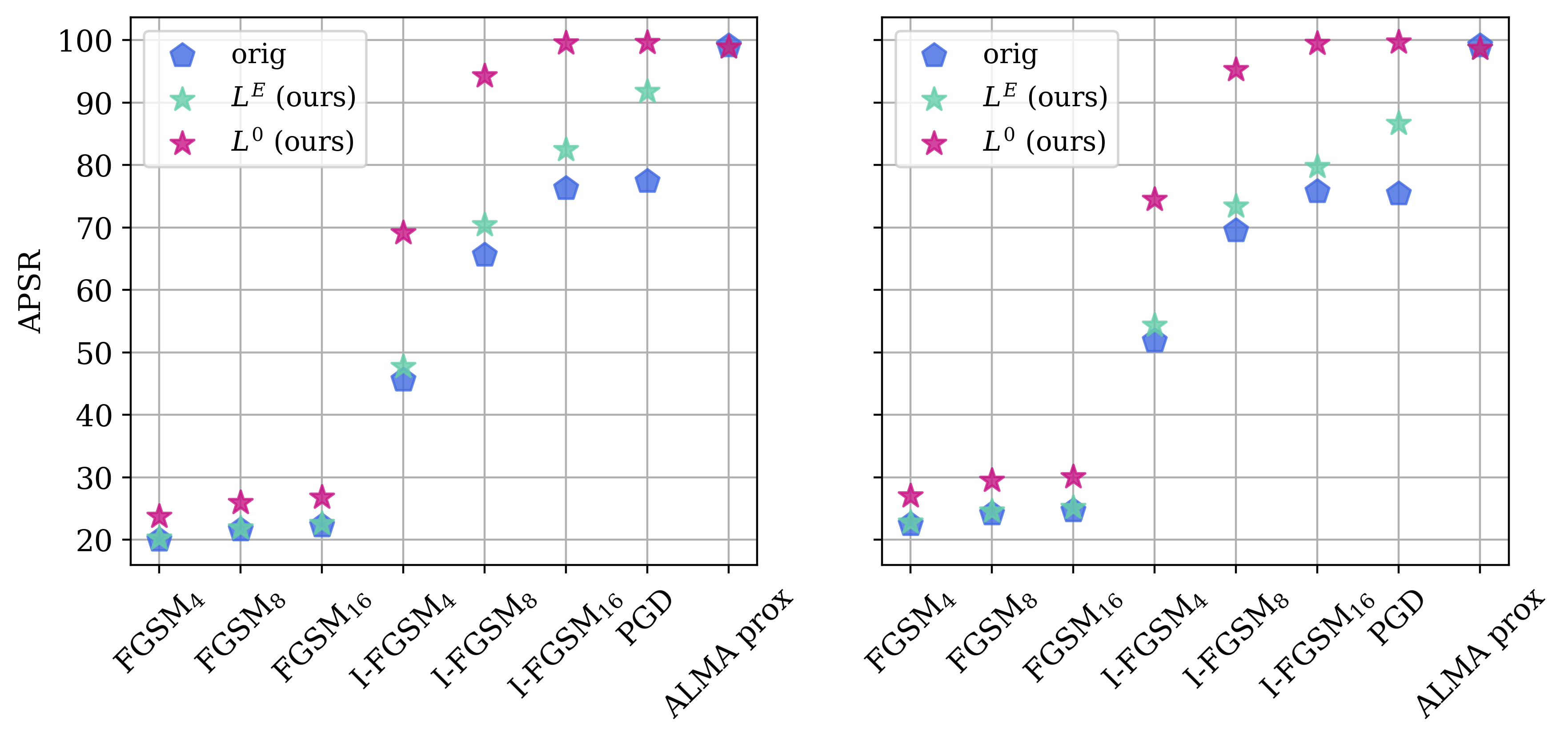}
    \caption{DeepLabv3+ (left) and PSPNet (right)}
    \end{subfigure}
    \begin{subfigure}[t]{.5\linewidth}
    \centering\includegraphics[width=0.99\textwidth]{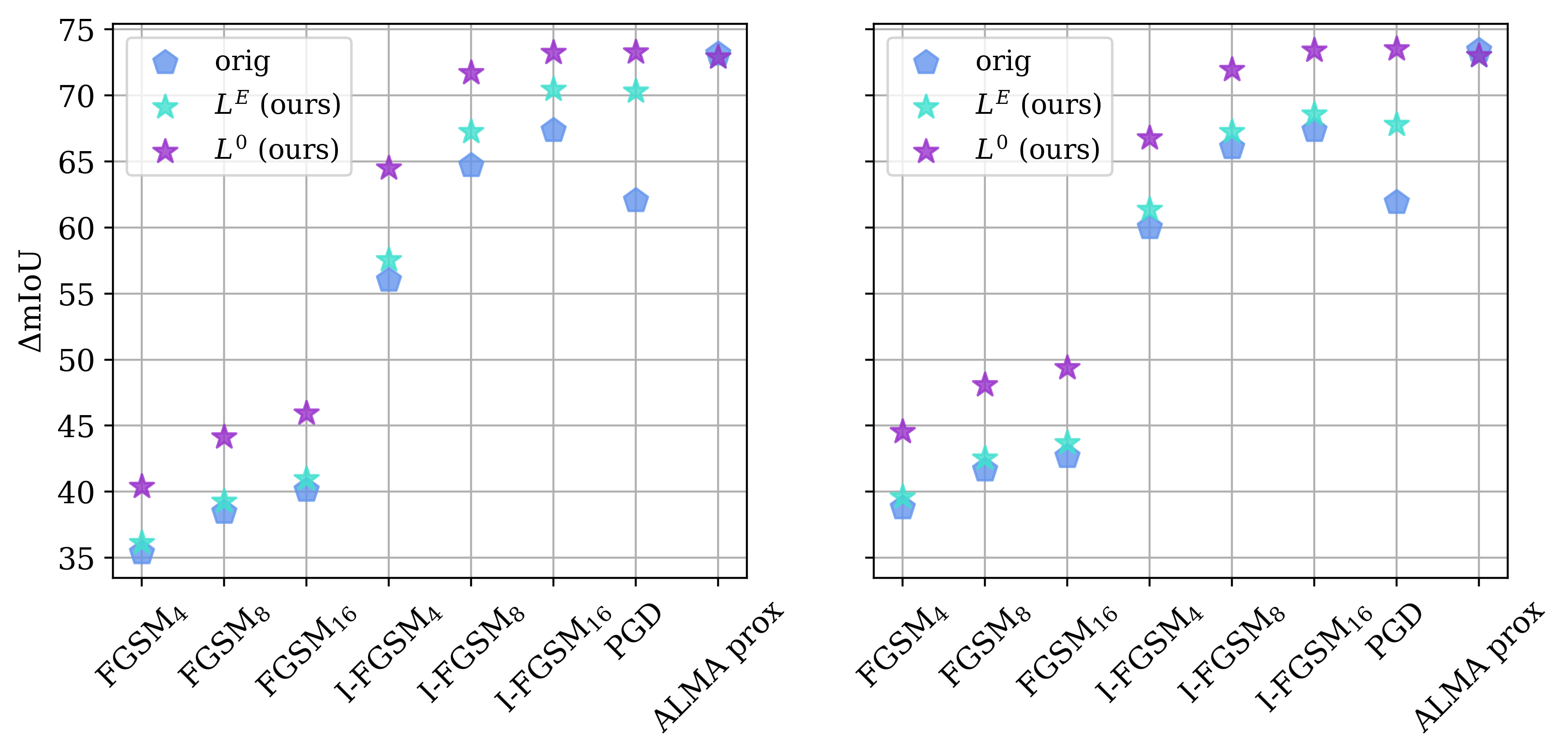}
    \caption{DeepLabv3+ (left) and PSPNet (right)}
    \end{subfigure}
    \caption{APSR (a) and $\Delta$mIoU (b) results for different attacks on two networks trained on the VOC dataset.}
    \label{fig:apsr_miou_voc}
\end{figure*}
A comparison between the attack performance of the original proposed attacks and the attacks resulting from replacing the original loss function with our incorporated uncertainty-weighted loss function, i.e., \cref{eq:loss_uw} with entropy as uncertainty measure and \cref{eq:loss_zero}, is given in \cref{fig:apsr_miou_cs} for the Cityscapes dataset and in \cref{fig:apsr_miou_voc} for the VOC dataset. 
We observe increased APSR and $\Delta$mIoU performance for larger magnitudes of perturbation for the non-iterative as well as the iterative FGSM attacks. Our approach clearly outperforms the original FGSM attack, as well in its simple as in its iterative version. 
Examples of segmentations stemming from the original I-FGSM as well as the uncertainty-weighed counterpart ($L^0$) are shown in \cref{fig:pred_example_diff} (b) and (f), respectively.
We obtain the largest performance boost for the PGD attack where the incorporation of the weighting leads up to $62.5$ percentage points (pp) higher APSR values. A closer inspection shows, that the original PGD attack performs poorly for the Cityscapes dataset. The reason for this is that using this attack, the same wrong class is often predicted for all pixels of an image. However, if the perturbation fails to do that, only a few pixels are predicted incorrectly. 
But with the uncertainty-weighted loss ($L^0$), almost all images are predicted incorrectly.
Alma prox is comparatively the strongest attack and achieves APSR values of over $99\%$. It is therefore difficult to improve the results any further, see for example \cref{fig:pred_example_diff} (c) and (g). The APSR values for the original attack and our uncertainty-based weighting scheme are very similar, although we can enhance the $\Delta$mIoU values for the Cityscapes dataset.
In general, the more extreme weighted loss function $L^0$ outperforms the entropy-weighted loss function $L^E$ for both datasets and investigated networks. This behavior is in the nature of the weighting manner, i.e., in $L^0$ pixels that are certainly incorrectly predicted are set to zero in the loss function and are therefore no longer considered, while in $L^E$ only a weaker weighting is applied. 
We also experimented with a combination of both loss functions which however did not increase the attack performance over using $L^0$ alone.

The runtimes are given in \cref{tab:run_times} for the different attackers.
\begin{table*}[t]
\centering
\scalebox{0.86}{
\begin{tabular}{ll c cccccccc}
\toprule
& & & FGSM$_4$ & FGSM$_8$ & FGSM$_{16}$ & I-FGSM$_4$ & I-FGSM$_8$ & I-FGSM$_{16}$ & PGD & ALMA prox \\
\cmidrule(r){4-11}
Cityscapes & Deep- & orig & $0.16$ & $0.16$ & $0.16$ & $0.74$ & $1.50$ & $3.01$ & $60.15$ & $64.75$ \\
& Labv3+ & ours &           $0.16$ & $0.16$ & $0.16$ & $0.77$ & $1.53$ & $3.07$ & $61.72$ & $66.80$ \\
\cmidrule(r){2-11}
& BiSe- & orig &            $0.14$ & $0.14$ & $0.14$ & $0.24$ & $0.36$ & $0.56$ & $9.81$ & $16.81$ \\
& Net & ours &              $0.15$ & $0.15$ & $0.15$ & $0.24$ & $0.36$ & $0.56$ & $10.05$ & $17.74$ \\
\cmidrule(r){1-11}
VOC & Deep- & orig & $0.08$ & $0.08$ & $0.08$ & $0.34$ & $0.69$ & $1.39$ & $27.41$ & $30.76$ \\
& Labv3+ & ours &    $0.08$ & $0.08$ & $0.08$ & $0.35$ & $0.71$ & $1.43$ & $28.16$ & $31.55$ \\
\cmidrule(r){2-11}
& PSP- & orig &      $0.08$ & $0.08$ & $0.08$ & $0.32$ & $0.64$ & $1.28$ & $26.21$ & $29.91$ \\
& Net & ours &       $0.09$ & $0.09$ & $0.09$ & $0.33$ & $0.65$ & $1.30$ & $26.96$ & $30.21$ \\
\bottomrule
\end{tabular} }
\caption{Runtimes (sec.\ per frame) for different adversarial attacks with original loss function in comparison to our uncertainty-weighted loss for both datasets and different networks.}
\label{tab:run_times}
\end{table*}
The results are averaged over the number of validation images and measured on a NVIDIA A40 GPU. The runtimes for the weighted loss functions are quite comparable and we quote the highest value here.
Incorporating our uncertainty-based weighting scheme into existing adversarial attack generation models increases the runtimes negligible but improves the performance greatly. 
Using our proposed weighting scheme, the I-FGSM$_{16}$ attack as well as the PGD attack attain similar performance values as the strong ALMA prox attack, but at lower runtimes. Only needing about $3$ to $5\%$  of  the runtime of ALMA prox, the advantage is especially drastic for the iterative FGSM. Note, the original attacks only achieve weaker performance.
%
%
\subsection{Comparison with CR attack}
The certified radius-guided approach is similar to our uncertainty-based weighting scheme as both methods weight the pixel-wise losses of the adversarial example generator to achieve a high attack success rate. We use the framework provided in the original paper \cite{Qu2023} and replace their CR weighting procedure with our uncertainty-based one to have a fair comparison of both methods. 
As shown in \cref{sec:results_loss}, the $L^0$ loss function performs best and is used in the following experiments. In \cref{tab:apsr_cr} (left), the numerical results are shown for the VOC dataset and the PSPNet. Note, 
the code for more datasets and models is not released by the authors up to now.
\begin{table}[t]
\centering
\scalebox{0.86}{
\begin{tabular}{l cc cc}
\toprule
& \multicolumn{2}{c}{APSR} & \multicolumn{2}{c}{$\Delta$mIoU} \\
& $\ell_{2}$ & $\ell_{\infty}$ & $\ell_{2}$ & $\ell_{\infty}$ \\
\cmidrule(r){2-3}\cmidrule(r){4-5}
orig & $59.92$ & $82.50$ & $61.36$ & $67.89$ \\
$L^0$ (ours) & $86.54$ & $99.69$ & $70.22$ & $73.46$ \\
\bottomrule
\end{tabular} }
\hspace{0.4ex}
\scalebox{0.86}{
\begin{tabular}{l c}
\toprule
clean & $3.64$ \\
orig & $5.18$ \\
ours & $3.70$ \\
\bottomrule
\end{tabular} }
\caption{APSR and $\Delta$mIoU results for the PSPNet applied to the VOC dataset comparing the CR framework with our approach (left). Corresponding runtimes in seconds per frame for the underlying attack without weighting scheme, the CR loss function as well as our approach (right).}
\label{tab:apsr_cr}
\end{table}
With up to $26.62$ pp higher APSR and $8.86$ pp higher $\Delta$mIoU values, our approach clearly outperforms the CR method for both parameter settings and evaluation metrics. 
Moreover, in \cref{tab:apsr_cr} (right) the runtimes for the adversarial attackers are given where ``clear'' means that the underlying attack is considered without any weighting scheme in the loss function. Our approach shows only a minimally extended runtime, while the CR method shows a runtime $1.4$ times larger. Thus, the proposed weighting scheme substantially improves over the CR based weighting in terms of performance and runtime. 
%
%
\subsection{Perturbation of a reduced number of pixels}
\begin{figure}
    \centering
    \includegraphics[width=0.43\textwidth]{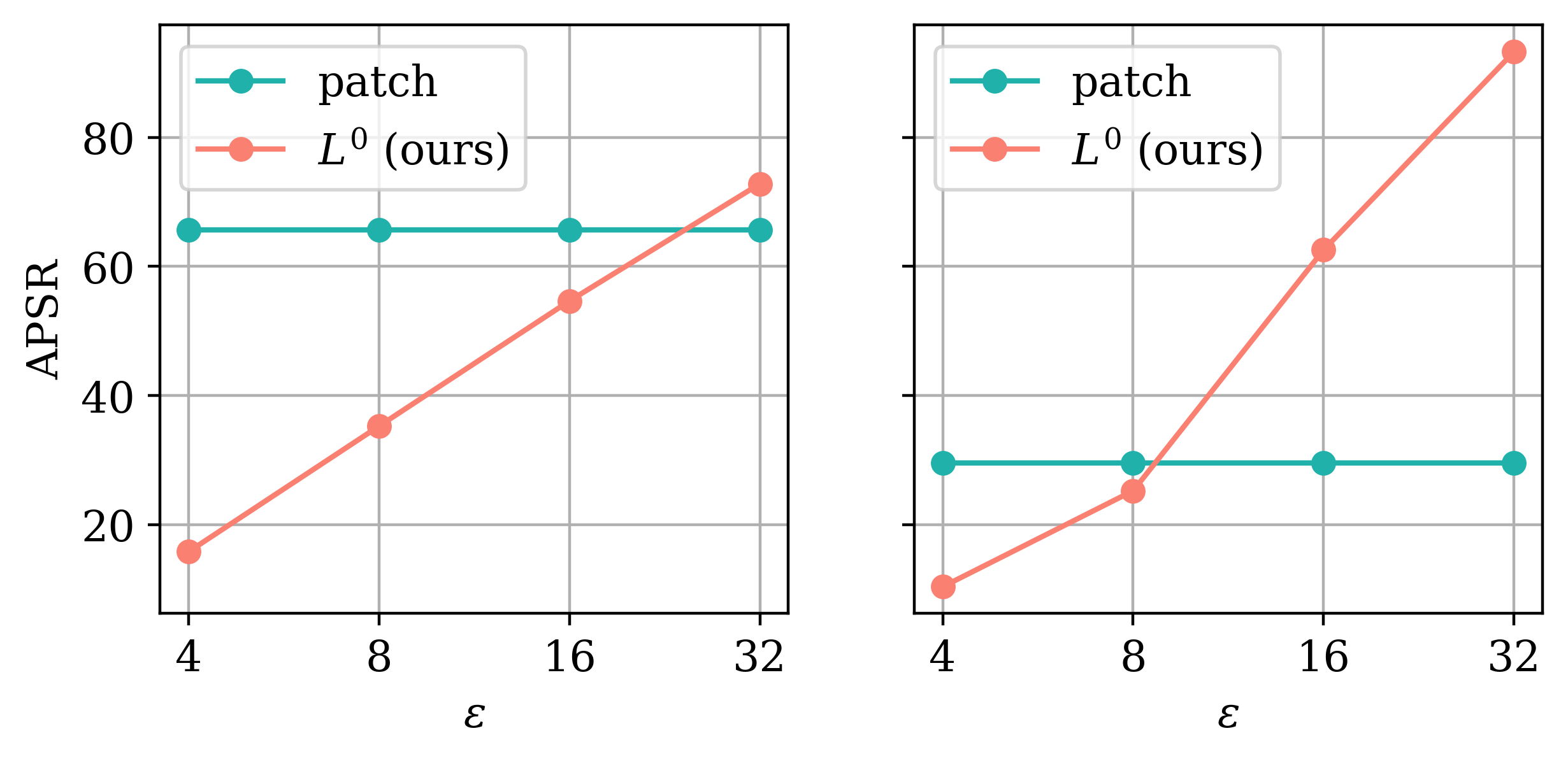}
    \caption{APSR results for the BiSeNet (left) and the DDRNet (right) applied to the Cityscapes dataset for a comparison between the patch attack and our approach (iterative FGSM attack with uncertainty-weighted loss applied to only a subset of pixels).}
    \label{fig:apsr_patch}
\end{figure}

Patch attacks disrupt a small rectangular region of the image aiming at prediction errors in a much larger region, i.e., the whole image.
We propose an alternative to the patch attack where also only a few pixels are perturbed. To this end, we choose randomly a subset of pixels to attack and apply our uncertainty-based loss function in combination with the iterative FGSM. Note, we disturb the same number of pixels like the considered patch method. Both methods are difficult to compare since the expectation over transformation-based attacking patch trains the attacker to successfully perturb the image over a range of transformations while our method perturbs only random pixels with the I-FGSM attack.
Our aim is to propose another way of attacking only a few pixels and achieving at the same time a high prediction damage. 
In \cref{fig:apsr_patch}, the performance results of the patch attack for the Cityscapes dataset are given in comparison to our approach using various magnitudes of perturbation for the I-FGSM attack. 
For the BiSeNet, we need a perturbation magnitude of $32$ to outperform the patch attack in terms of APSR, while for the DDRNet a magnitude of $16$ is sufficient.
In \cref{fig:pred_example_diff} (d) and (h), a qualitative result of the patch attack and our approach is shown. Both attacks perturb the prediction in different ways, i.e., the patch attack targets the upper part of the image while our approach focuses on the lower part. 
This observation is not specific to the shown example but holds more generally for segmentations of the discussed adversarial examples. 
%
%
%
\section{Related work}\label{sec:rel_work}
The only similar work to our uncertainty-based weighting scheme is the certified radius-guided approach proposed in \cite{Qu2023} focusing on the attack of pixels with relatively smaller certified radii. While we use different uncertainty measures that are simple to compute or set values in the loss function to zero, the calculation of the certified radius for each pixel produces high computational overhead which is reflected in the runtimes of the method. In addition to the more expensive calculation and longer runtimes, the attack performance is also worse compared to our method. 

The patch attack \cite{Nesti2022} perturbs only a few pixels of an image, i.e., a rectangular region of fixed size. The expectation over transformation-based method trains the attacker to successfully perturb the image over a range of transformations. In contrast, our approach also perturbs only a small number of pixels (exactly the same number as the patch attack) but uses the I-FGSM procedure with uncertainty-weighted loss and targets only a specific image. Therefore, both methods are not comparable with each other, rather, they demonstrate two different ways to create a high degree of attack damage to the image while perturbing only a few pixels. 
\Cref{fig:pred_example_diff} shows that both attacks focus on different regions, so combining both attacks could be interesting. 
%
%
%
\subsection*{Potential negative societal impact}
Adversarial attacks are generally considered to be malicious, as they can compromise the security of neural networks and rapidly degrade performance. However, the development and especially the free availability are of highest interest to develop detection and defense methods to prevent attacks.
%
%
%
\section{Conclusion and outlook}\label{sec:conc}
In this work, we proposed an uncertainty-based weighting scheme which can be incorporated into the loss function of any untargeted attack on semantic segmentation models that is composed out of pixel-wise attacks. We exploited the correlation between uncertainty measures and erroneous predictions to strongly degrade the prediction performance of neural networks. The expectations for any attack are low runtimes and computational effort while at the same time having powerful perturbation effects. Our approach can be applied to any attack with minimal computational overhead compared to the original attack, but results in significantly enhanced perturbation performance. We achieved attack pixel success rate values of up to $99.82\%$ across different network architectures and datasets. Moreover, we presented a method which attacks only a subset of the image pixels, similar to the patch attack, but also leading to an erroneous prediction of big parts of the image.

As a further improvement, we plan to develop an uncertainty-weighted loss function for targeted adversarial attacks as our approach is limited to untargeted attacks. 
%
\section*{Acknowledgement}
This work is supported by the Ministry of Culture and Science of the German state of North Rhine-Westphalia as part of the KI-Starter research funding program and by the Deutsche Forschungsgemeinschaft (DFG, German Research Foundation) under Germany’s Excellence Strategy – EXC-2092 CASA – 390781972.

{\small
\bibliographystyle{ieee_fullname}
\bibliography{egbib}
}

\end{document}